# The Application of Active Query K-Means in Text Classification

Yukun Jiang
*Department of Computer Science*
*New York University*
New York, 10003, United States
jy2363@nyu.edu

*Abstract*—Active learning is a state-of-art machine learning approach to deal with an abundance of unlabeled data. In the field of Natural Language Processing, typically it is costly and time-consuming to have all the data annotated. This inefficiency inspires out application of active learning in text classification. Traditional unsupervised k-means clustering is first modified into a semi-supervised version in this research. Then, a novel attempt is applied to extend the algorithm into active learning scenario with Penalized Min-Max-selection, so as to make limited queries that yield more stable initial centroids. This method utilizes both the interactive query results from users and the underlying distance representation. After tested on a Chinese news dataset, it shows a consistent increase in accuracy while lowering the cost in training.

## I. INTRODUCTION

As a state-of-art approach to interactively querying a small amount of labeled data in combination with a large amount of unlabeled data during training process, active learning has been applied in many fields, but a couple of shortcomings have also gradually been uncovered, particularly in Natural Language Processing with the ignorance of massive unlabeled data, which inspires us to reconsider the possibility of applying active learning approach in Text Classification by using experiment data collected from China News.

The K-means clustering algorithm is one of the most fundamental unsupervised machine learning algorithms, which can split documents into pre-determined number of clusters, whose structures, however, may not live up to human's primary expectation. In addition, due to the intrinsic randomness, it could take quite a long time for the algorithm to converge. These situations motivate this study first to implement semi-supervised k-means and then transform it into active query k-means to solve the two problems aforementioned. In particular, this study plans to modify the pure distance-based Min-Max selection strategy into a penalized form by active learning to fully utilize the query. Some related algorithms have already been investigated including Fuzzy-CMeans [1], seed-based K-means [2] and semi-supervised Density-Based Clustering [3].

## II. RELATED WORKS

The inherent defect of unsupervised clustering algorithms is in their structures: they are found to possibly contradict or miss the desired structure suitable for the underlying applications [4], which is a common problem for NLP text classification or topic modeling applications when using clustering or Latent Dirichlet Allocation (LDA). For improvement, Sahoo indicates that relatively high accuracy can be achieved with the assistance of 10% labeled information in medical document classification.

The use of an excessive amount of RAM memory by K-means related algorithm presents a practical problem to many practitioners. Senuma [5] proposed to adopt a hash function for dimension reduction, because hashing is likely to provide an efficient way of processing high dimensional vectors, which can make a huge difference for this study' algorithms in training processes, given the diversity and thus high dimensional vectors of Chinese words and phrases.

Viet-Vu Vu and his teams [6] proposed a pioneering active learning variant of K-means clustering. Specifically, in usual semi supervised K-means, the initial centroids are computed by using a small portion of labeled data provided. But the performance may vary depending on whether the labeled data provided is balanced or close to the real centroids. The proposed active learning variant of K-means seeks to pick the portion of labeled data that favor the coverage of the whole dataset. Their pure distance-based Min-Max Selection algorithm purposed is inspiring and is the prototype of our study.

## III. PREPARATION

### A. Dataset

The dataset for this study is collected from China News, which covers 7 classes of news: "Domestic", "HK- Taiwan", "International", "Economics", "Art", "Celebrity" and "Sports". The dataset contains 1.3 million entries of short news for training and 0.11 million for testing, both with answer keys and are roughly well-balanced, with each class accounting for about 14%. Owning to training time constraints, 4000 news and 500 news for train and test respectively are applied respectively. Furthermore, the trainset is splitted into 80% for training and 20% for validation.

## B. Data Preprocessing

Since the dataset is in Chinese, there are no spaces between words as a token splitter. Therefore, this study performs the following preprocessing steps:

1) **tokenization**: the python package "jieba" is applied with common stop-word list and punctuation lists for Chinese word to tokenize each news piece.

2) **lemmatization**: some traditional Chinese characters present in the datset are transformed into simplified Chinese characters for consistency and lower complexity.

3) **vectorization**: after TF-IDF vectorization, each news piece's feature dimension is 15283 long, which would for sure present computational difficulty in the training process. Thus, dimension reduction is necessary.

4) **dimension-reduction**: feature Hashing technique as suggested by Hajime Senuma's article using hashing function [5] was first applied. However, the result is found not satisfactory since many unrelated words are hashed into the same entry, making it impossible for the system to tell apart the two news pieces. For example, given the following two sentences:

1. Monkeys love banana. (Animal category)

2. Virus breaks out. (Pandemic category)

Then the vector would be 6-long, corresponding to index1: monkey, index2: love, index3: banana, index4: virus, index5: break and index6: out.

The vector for Sentence 1 is [1,1,1,0,0,0], for Sentence 2 [0,0,0,1,1,1]. If we were to reduce dimension by 50%, namely making the vector to be 3-long, we would do "modulus 3" operation. After taking the modulus, both vectors for Sentence 1 and Sentence 2 are [1,1,1]. Clearly any system could not tell these two news pieces apart because of the same vectorization.

Therefore, the principal component analysis (PCA) is applied to effectively reduce the dimension. The statistical explained variation measures the proportion to which a model accounts for the dispersion of the given data set. To balance between dimensionality and information retainment, this study chooses to adopt a 5% PCA level with an explained variance 41%.

TABLE I. PCA STATISTICS

| PCA level | 1% | 5% | 10% | 20% |
|---|---|---|---|---|
| Explained variation | 19% | 41% | 56% | 68% |

## IV. EVALUATION MEASURE

### A. Performance Measure

Since this is a text classification task, the length of system output is always consistent with the length of answer keys. Therefore, accuracy is used as the evaluation metric, namely $$\boldsymbol{ccuracy} = \frac{correct\ classifications}{total\ num\ of\ news}.$$

### B. Uniformity Measure

In order to find out how "uniformly distributed" the initialization of the dataset picked is, we use Gini-Index as measure. The higher the Gini-Index is, the more uniform the dataset is. The definition is as follows:

**Definition 1**. *For a dataset containing k different classes with each class i appearing with probability $p_i$ such that $\sum_{i=1}^{k} p_i = 1$, and the Gini-Index is $\sum_{i=1}^{k} p_i(1 - p_i) = 1 - \sum_{i=1}^{k} p_i^2$*

**Theorem 1**. *For a dataset containing k different classes with each class i appearing with probability $p_i$, the Gini-Index is maximized when $p_i = \frac{1}{k}, \forall i \in \{1, \cdots, k\}$*

Thus, in our dataset which contains 7 categories of news, the theoretical upper bound is when each class appears with probability $\frac{1}{7}$, yielding ideal Gini-Index $= 1 - \sum_{i=1}^{7} \left(\frac{1}{7}\right)^2 = 0.857$.

### C. Baselines & Upper Bound

Two baselines and a single upper bound are adopted to be compared with two systems (semi-supervised k-means & active query k-means)

**Baseline1**: Baseline1 is designed for semi-supervised k-means with random initialization. The intuitive strategy in text classification is to predict the most frequent class. As our dataset is almost uniformly distributed and well-balanced, the baseline in this case would be $\frac{1}{7} \approx 14\%$. Therefore, the algorithm should achieve more than 14% accuracy.

**Baseline2**: Baseline2 is designed for active query k-means with Penalized Min-Max-selection. Since this algorithm makes queries on the most uncertain cases during initialization stage, it is expected to outperform the randomly initialized version semi-k-means since it should enjoy a more "uniformly distribution" thus getting more reliable initial clusters of centriods.

**Upperbound**: Given that both semi-supervised learning and active learning strategies are not utilizing all the information given by the training dataset, it's natural that the performance would be inferior to some fully-supervised learning algorithms. We propose to use multiclass Support Vector Machine (SVM) to be the upper bound for performance since SVM is one of the most traditional and widely used ML algorithms in text classification. After tuning the hyperparameters, the best setting of SVM algorithm with hinge loss function and linear basis gives 70.42% accuracy on the test dataset.

## V. ALGORITHMS

### A. Metric

The metric used is Euclidean Distance. Specifically, let $\vec{p}$ and $\vec{q}$ be two vectors of dimension n, then the Euclidean Distance between them is $d(\vec{p}, \vec{q}) = \sqrt{\sum_{i=1}^{n}(p_i - q_i)^2}$. And in active query scenarios, this metric is modified into a penalized form.

## B. Unsupervised K-means

The starting algorithm is built upon the traditional unsupervised k-means, also referred to as Lloyd's algorithm [8]. Given a dataset D and a predetermined cluster number k, the algorithm seeks to partition the dataset into k clusters in which each datapoint belongs to the cluster with the nearest mean (the so-called centroid), serving as a prototype of this cluster. The training process mainly involves continuously updating the centroids and reassigning datapoints to their nearest clusters with updated centroids. The time complexity for this algorithm is $O(n^2)$. A detailed algorithm and a small demo are given as follows:

| ALGORITHM 1: Unsupervised K-means |
|---|
| Input: Dataset D, number of clusters k |
| Output: D partitioned into k clusters |
| 1: arbitrarily choose k datapoints from D as initial centroids |
| 2: repeat |
| 3: reassign each datapoint to nearest clusters |
| 4: compute mean of datapoints in each cluster |
| 5: update each cluster's centroid using the mean |
| 6: until cluster convergence |

The small demo below indicates the procedure of the unsupervised K-means. There are 3 cluster of Gaussian random samples generated and unlabeled. After applying the algorithm till convergence, they are correctly partitioned into 3 clusters with each one's centroid highlighted (by red star) in the graph.

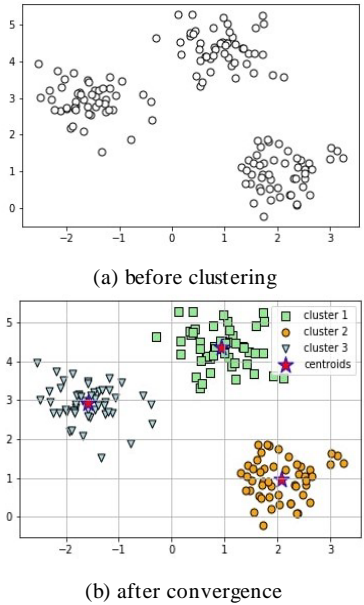

(a) before clustering

(b) after convergence

Fig. 1. Visualization of unsupervised K-means.

## C. Semi K-Means

The main difference between the semi K-Means clustering and the traditional unsupervised K-means clustering lies in the initialization stage. Instead of just randomly picking some points as the centroid for each cluster, semi K-Means algorithm explore the small portion of labeled data and use that additional information to build the initial clusters of centroids. In the ideal scenarios, if each actual cluster has some datapoints prelabelled, then the initial centroids would already be roughly correct in the location distribution in Euclidean space. Of course, there could be bad cases that the distribution of initially labeled dataset is skewed. A detailed algorithm and a demo of initialization stage are given as follows:

| ALGORITHM 2: Semi K-means |
|---|
| Input: Dataset U(unlabeled), L(labeled), number of clusters k |
| Output: D∪L partitioned into k clusters |
| 1: build initial centroid for each cluster using L |
| 2: repeat |
| 3: reassign each datapoint to nearest clusters |
| 4: compute mean of datapoints in each cluster |
| 5: update each cluster's centroid using the mean |
| 6: until cluster convergence |

The small demo below presents how the algorithm uses the initially labeled data to do initialization. In the demo there are 8 points labeled, 2 from class0, 4 from class1 and the rest 2 from class2. The algorithm uses them to compute the initial centroids (the red stars in the graph) by taking the mean and then start the updating iteration until convergence.

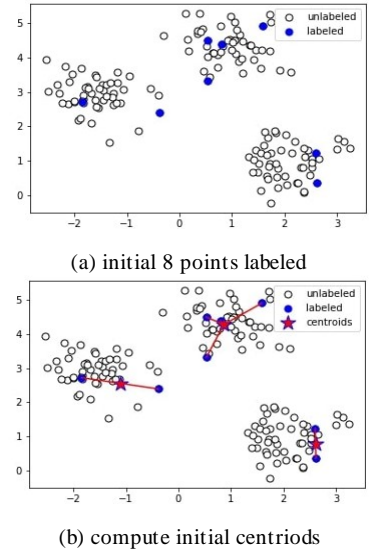

(a) initial 8 points labeled

(b) compute initial centroids

Fig. 2. Sample initialization of semi K-Means.

## D. Active Query K-means

Active Query K-means is a variant of semi K-Means used to interact with expert annotators in selecting the initial seeds. In order to make the best use of queries, it no longer randomly picks a proportion of labeled data. The query strategy this study implements is Penalized Min-Max Selection. The basic underlying assumption behind this strategy is that, the more scattered the seeds are, the more likely they uniformly cover the whole dataset. Besides, the label queried should be fully exploited to pick datapoint from low-frequency classes. Mathematically, given the whole trainset X and selected seeds set Y containing $\vec{y_i}$, $i \in \{1,2,\cdots,n\}$, the next point $\vec{y_{n+1}}$ to pick is decided as follows:

$$\vec{y}_{n+1} = \arg\max_{\vec{x} \in X} \left( \min_{\vec{y_i} \in Y} d^{\text{penalized}}\left(\vec{y_i}, \vec{x}\right) \right) \quad (1)$$

and

$$d^{\text{penalized}}\left(\vec{y_i}, \vec{x}\right) = \Phi\left(k_{y_i}\right) d\left(\vec{y_i}, \vec{x}\right) \quad (2)$$

where $k_{y_i}$ is the number of datapoint already picked in Y having the same label as $y_i$ and $\Phi()$ is the penalize function such that it is a non-negative and decreasing function. In interpretation, the next point $\vec{y_{n+1}}$ should be picked far away from each already-picked point $\vec{y_i}$, $i \in \{1,2,\cdots,n\}$, so that it's likely that the label of $\vec{y_{n+1}}$ could be in a class that has not previously been selected or is currently under-represented. A detailed algorithm and a demo of initial queries are as follows:

| ALGORITHM 3: Active Query K-means |
|---|
| Input: Data set X, number of queries k |
| Output: set of seeds Y |
| 1: Y = ∅ |
| 2: randomly pick a point $\vec{y_1}$ into Y from X |
| 3: query about the label of $\vec{y_1}$ |
| 4: UnSelected = X\ $\vec{y_1}$ |
| 5: while Y.size < k do |
| 6: $\vec{next}$ = Penalized Min-Max(Y, UnSelected) |
| 7: query about the label of $\vec{next}$ |
| 8: Y = Y ∪ $\vec{next}$ |
| 9: UnSelected = UnSelected \ $\vec{next}$ |
| 10: end while |
| 11: return Y |

| ALGORITHM 4: Penalized Min-Max Selection |
|---|
| Input: Y, UnSelected |
| Output: $\vec{next}$ |
| 1: for all $x_i \in$ UnSelected do |
| 2: $d^{\text{penalized}}_{\vec{x_i}} = \min_{\vec{y} \in Y} d^{\text{penalized}}\left(\vec{y}, \vec{x_i}\right)$ |
| 3: end for |
| 4: $\vec{next} = \arg\max_i \left( d^{\text{penalized}}_{\vec{x_i}} \right)$ |
| 5: return $\vec{next}$ |

The small demo below illustrates how the Penalized Min-Max Selection strategy picks the initial seeds. There are 200 points from three different cluster. Allowing 10 queries, the initial seeds are quite scattered and uniform over the whole dataset.

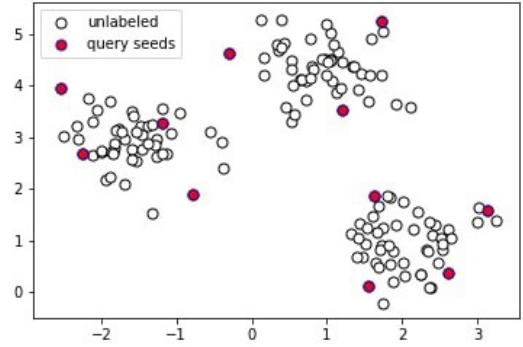

Fig. 3. Initial 10 queries using active query.

## VI. EXPERIMENTS

### A. Semi K-Means

Experiments are conducted to evaluate the performance of semi K-Means algorithm on the test set. First a small portion of data are labeled and fed into the model to compute the initial centriods. And then the algorithm runs on the training set to finalize each cluster's position until convergence. Afterwards, it is applied to the test dataset to evaluate the performance. The experiments are conducted in a 10-fold manner to take the average, in order to overcome the effect of randomness from initialization. The portion of initial labeled data ranges from 1 percent to 10 percent to observe the scalability and stability of the algorithm.

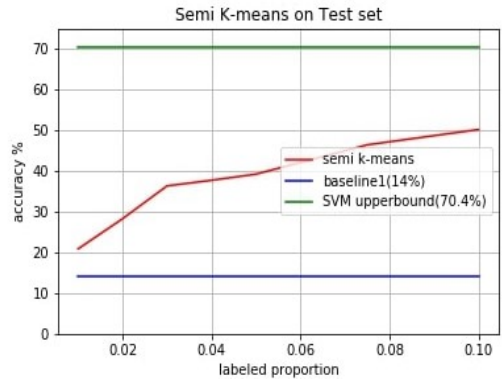

Fig. 4. Experiment result of semi K-Means.

TABLE II.  GINI-INDEX

| Labeled proportion | 1% | 2% | 3% | 4% | 5% | 7.5% | 10% |
|---|---|---|---|---|---|---|---|
| Accuracy | 20.87% | 28.28% | 36.34% | 37.71% | 39.22% | 46.41% | 50.16% |
| Gini-Index | 0.8105 | 0.8320 | 0.8010 | 0.7850 | 0.8425 | 0.8350 | 0.8225 |

From the graph and table, clearly Semi K-means algorithm beats this study's baseline1(14%) of choosing the most likely class and is bounded above by SVM upperbound(70.42%). The highest accuracy when using 10% label proportion is 50.16%, with a 20.26% gap compared to SVM. As the labeled proportion increases, the accuracy first increases dra- matically and then gradually slow down approaching asymp- totic horizontal line, which indicates that even higher label proportion might not lead to higher accuracy.

Also the Gini-Index reveals the intrinsic randomness of this algorithm. The Gini-Index does not necessarily increase and actually fluctuates heavily, even if the experiments use 10-fold validation and the proportion of labeled data in- creases. Based on the above analysis, it is conjectured that a more uniform initial seeds with higher Gini-Index could lead to better performance and accuracy. Active Query K-means plays an important role to answer the question.

### B. Active Query K-means

Similar Experiments are conducted by using Active Query K-means on the test dataset using penalized min-max selection. In the experiments, we tried different penalize functions including $\Phi(x) = \frac{1}{e^x}, \frac{1}{\log x}, \frac{1}{x}, \frac{1}{x^2}, \frac{1}{\sqrt{x}}$. Among these penalize functions, $\Phi(x) = \frac{1}{\sqrt{x}}$ has the best performance and thus we adopt it in the following experiment. Given an initial query ratio, a small subset of train dataset is selected and their labelled are provided. Using this subset as initial seeds, the algorithm runs on the whole train dataset and then is tested on the test dataset. Since first point in the query sequence is taken randomly, the experiment is conducted 10-fold for each query ratio with the average which can minimize the effects of randomness. The query ratio ranges from 1 percent to 10 percent to observe if adequate queries yield more uniform initial seeds and thus a better performance.

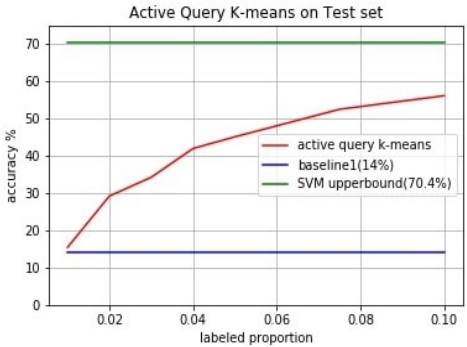

Fig. 5. Experiment of active query K-means.

TABLE III. GINI INDEX

| Labeled proportion | 1% | 2% | 3% | 4% | 5% | 7.5% | 10% |
|---|---|---|---|---|---|---|---|
| Accuracy | 15.48% | 29.21% | 34.24% | 41.96% | 45.08% | 52.46% | 56.09% |
| Gini-Index | 0.8004 | 0.8386 | 0.8406 | 0.8444 | 0.8487 | 0.8500 | 0.8527 |

From the table and plot, Active Query K-means does beat baseline1(14%) of choosing the most likely class and also is bounded above by SVM Upperbound (70.42%). The highest accuracy using 10% query ratio is 56.09%, having a 14.33% gap compared with SVM. What's more, unlike semi k-means whose graph approaches asymptotic horizontal line, it is observed that the accuracy graph for active query k-means is slightly concave down, meaning that if higher query ratio provided, accuracy could continue to increase substantially.

### C. Comparison

By using Penalized Min-Max Selection in seeds initialization, initial seeds are more uniformly covering all the classes in the dataset. Consequently, the initial centriods would be closer to the "real" centriods for each class according to probability distribution.

In the graph, the barplot is the accuracy on Test Set and the lineplot is the Gini Index for the initial seeds. After 3% labeled/query ratio, Active Query K-means always surpasses Semi K-means quite significantly. With 10% query ratio, the accuracy gap between them is 6%.

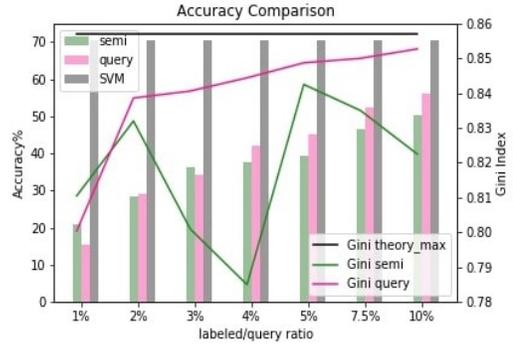

Fig. 6. Comparison between semi and active query K-means.

GIni Index plot provides insights into the reasons behind. By Theorem, in our case of 7 classes, the highest theoretical upper bound for Gini Index is 0.857 (the black horizontal line). The Gini Index for Active Query increases steadily as the query ratio increases, while the Gini Index for Semi fluctuates. In another word, the more queries allowed, the more accurate initial centroids would be in Active Query algorithm, whereas in Semi algorithm the initial centroids could still be skewed. This could not only explain the difference in accuracy performance, but also interpret scalability that if higher labeled/query ratio allowed, we expect the performance for active query algorithm to increase more sharply than of the semi algorithm.

## VII. ANALYSIS

### A. Error Statistic

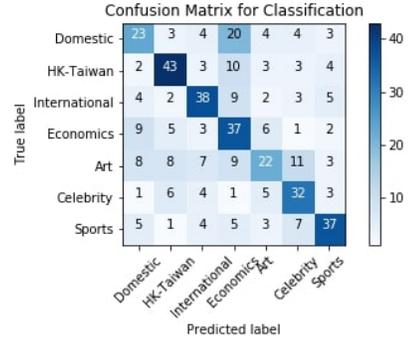

Fig. 7. Confusion matrix for test.

The confusion matrix and table statistics clarify how and where classification error occurs. Two of the most frequent mis-classifications are to classify a piece of Domestic news as Economics and a piece of Art news as Celebrity. Further inspection of dataset reveals that some news is inherently mix-classed and thus hard to distinguish. There are two typical examples to support this point:

1. 国家再增加 16 亿元云南鲁甸地震应急救灾资金
(The state will increase its emergency relief fund for the earthquake by 1.6 billion yuan)
True label: **Domestic** Predict label: **Economics**

2. 迈克尔·苏立文著作《中国艺术史》出版
(Michael Sullivan's book *The History of Chinese Art* has been published)
True label: **Art** Predict label: **Celebrity**

Fig. 8. Error Example.

The two pieces of news are difficult to classify even for human annotators, since either the true label or predicted label seems quite reasonable. Thus, the existence of such "mixed" news in dataset would certainly decrease the system's performance.

*B. Information Loss*

There are two major sources of information loss in our system that might result in performance decrease.

1) During the data preprocessing stage, all the traditional Chinese characters that appear in our dataset are transformed into their simplified form of chinese. However, thanks to our other colleagues, we notice that some characters might carry different meanings in traditional and simplified chinese context. And even when they do carry the same meanings, they might be used in different contexts according to whether in traditional or simplified chinese. Such difference might be too subtle to be spotted one by one. Possible solutions to this subtlety are discussed in details in David's articles [7].

2) In the PCA dimension reduction stage, we shrink the TF-IDF vectors' dimension to 5% of the original ones with explained variation 41%. Some information is lost in exchange of smaller dimensionality [8], but a faster training process is delivered. Instead of shrinking the dimension from the beginning, a model that generates sparsity might be considered. The $L_1$-regularized *lasso* variants give a large class of such sparse models [9].

TABLE IV. PERFORMANCE

| Class | Domestic | HK-Taiwan | International | Economics | Art | Celebrity | Sports | Average |
|---|---|---|---|---|---|---|---|---|
| Precision | 44.23 | 63.23 | 60.31 | 40.66 | 48.88 | 52.45 | 64.91 | 51.16 |
| Recall | 37.70 | 63.23 | 60.31 | 58.72 | 32.35 | 61.54 | 59.68 | 54.30 |
| F1 | 40.47 | 63.23 | 60.31 | 48.05 | 39.93 | 56.63 | 62.19 | 52.68 |

VIII. FUTURE WORK

For future explorations, with sufficient time and access to more powerful machines, we intend to train and test the system on the whole dataset which is 1.3 million entries large instead of a fraction of it as studied now. By feeding the system with more training data, we expect the performance could increase even more. Also, more penalize functions could be experimented on.

Besides, we also plan to improve the way of computing distance and taking Min-Max, because under current algorithm the time complexity is nonlinear and as dataset gets larger, the time needed to compute distance and take Min-Max would increase sharply. By the theory of convex optimization, by solving the duality problem [10], we expect to enjoy the cost and time for training to decrease.

Furthermore, the study should also include a couple of skewed and imbalanced dataset to explore the comprehensive robustness of the algorithms. And following Castro and Nowak's article [11], the convergence and error bounds of this algorithm could be explored in more details mathematically

IX. CONCLUSIONS

This study introduces a novel active query strategy — Penalized Min-Max Selection — in the active K-means clustering algorithm. As a modification of pure distance-based Min-Max-Selection and other label-based active learning strategies, this query strategy seeks to exploit the query labels and the distance representation simultaneously. As shown in the experimental result, the new strategy surpasses the random seed-based K-means algorithm by 6% if allowing 10% queries; and the curvature of accuracy plots show that a higher accuracy would be attainable if a higher query ratio is allowed. In future studies the algorithm should be experimented on larger and imbalanced dataset, so as to test its robustness and full strength.